\documentclass[10pt,twocolumn,letterpaper]{article}
\pdfoutput=1
\usepackage{iccv}
\usepackage{times}
\usepackage{epsfig}
\usepackage{graphicx}
\usepackage{amsmath}
\usepackage{amssymb}

\usepackage{graphicx}
\usepackage{amsmath}
\usepackage{amssymb}
\usepackage{booktabs}
\usepackage{enumitem}

\usepackage[ruled]{algorithm2e}
\usepackage{textcomp}
\usepackage{siunitx}
\usepackage[table,x11names]{xcolor}

\usepackage[pagebackref=true,breaklinks=true,letterpaper=true,colorlinks,bookmarks=false]{hyperref}

\newcommand{\X}{\mathbf{X}}
\newcommand{\Y}{\mathbf{Y}}
\newcommand{\x}{\mathbf{x}}
\newcommand{\y}{\mathbf{y}}
\newcommand{\z}{\mathbf{z}}
\newcommand{\name}{{\color{black} RISE}} 

\iccvfinalcopy 

\def\iccvPaperID{9676} 

\ificcvfinal\fi

\newcommand\blfootnote[1]{%
  \begingroup
  \renewcommand\thefootnote{}\footnote{#1}%
  \addtocounter{footnote}{-1}%
  \endgroup
}

\begin{document}

\title{
A Sentence Speaks a Thousand Images:\\
Domain Generalization through Distilling CLIP with Language Guidance
}

\author{
Zeyi Huang$^1$ \hspace{0.6cm} Andy Zhou$^2$ \hspace{0.6cm} Zijian Lin$^3$ \hspace{0.6cm} Mu Cai$^1$ \hspace{0.6cm} Haohan Wang$^{2\dagger}$ \hspace{0.6cm} Yong Jae Lee$^{1\dagger}$ 
\\
{$^1$University of Wisconsin-Madison \hspace{0.3cm} $^2$University of Illinois Urbana-Champaign} \hspace{0.3cm} $^3$Imperial College London\\
\tt\small{ \{zeyihuang, mucai, yongjaelee\}}@cs.wisc.edu  \tt\small{ \{andyz3,haohanw\}}@illinois.edu  \tt\small{ z.ling22}@imperial.ac.uk\\
}

\maketitle
\ificcvfinal\thispagestyle{empty}\fi

\begin{abstract}
Domain generalization studies the problem of training a model with samples from several domains (or distributions)
and then testing the model with samples from a new, unseen domain.
In this paper, 
we propose a novel approach for domain generalization that leverages recent advances in large vision-language models, specifically a CLIP teacher model, to train a smaller model that generalizes to unseen domains. The key technical contribution is a new type of regularization that requires the student's learned image representations to be close to the teacher's learned text representations obtained from encoding the corresponding text descriptions of images. 
We introduce two designs of the loss function, absolute and relative distance, which provide specific guidance on how the training process of the student model should be regularized. 
We evaluate our proposed method,
dubbed \name{} (Regularized Invariance with Semantic Embeddings),
on various benchmark datasets, and show that it outperforms several state-of-the-art domain generalization methods. 
To our knowledge, our work is the first to leverage knowledge distillation using a large vision-language model for domain generalization. By incorporating text-based information, \name{} improves the generalization capability of machine learning models.

\end{abstract}

\section{Introduction}
\blfootnote{$^\dagger$ equal advising. Code is available at \href{https://github.com/OoDBag/RISE}{\color{black}{github.com/OoDBag/RISE}}}
\begin{figure}
    \centering
    \includegraphics[width=10cm,height=5.7cm,keepaspectratio]{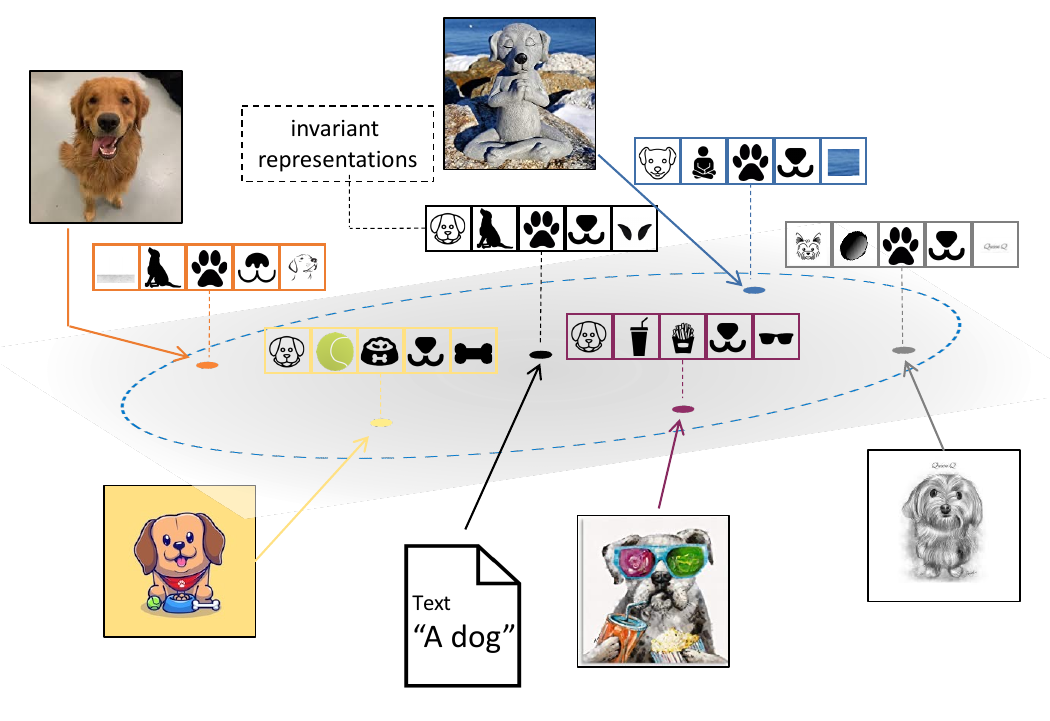}
    \caption{The key intuition behind our argument. While images can capture more details, text can directly summarize the core concept to represent the object of interest.}
    \label{fig:open}
\end{figure}

An image is worth a thousand words, 
indeed,  because of its power to convey a wealth of information through its visual details.
However,  a well-written sentence, on the other hand, has the power to concisely capture the essential information that is common to many 
different images. 
By describing a scene with a few carefully chosen words, 
a writer can create a mental image in the reader's mind that conveys the essence of what is being depicted. 
This perspective is particularly useful when 
communicating information efficiently, 
or when emphasizing a specific scene aspect without getting bogged down in extraneous details.
Thus, 
we suggest that 
a sentence speaks a thousand images. 

Essential semantic information delivered by an image plays a pivotal role in helping models generalize to shifted distributions, whereas other detailed information (e.g., in the background not relevant to the main object) captured in images may not be as effective for this purpose. The study of domain generalization~\cite{muandet2013domain} investigates the problem of training a model with samples from several domains (or distributions) and then testing the model with samples from a new, unseen domain. The training domains are commonly referred to as source domains, and the test domain is referred to as the target domain. Previous studies have identified a challenge in training effective domain generalization models due to the models' tendency to learn domain-specific features \cite{geirhos2018imagenet}. Consequently, numerous works have focused on regularizing the models to learn representations that are invariant to domain-specific features \cite{li2018domain,li2018deep,zhao2020domain,wang2016select,motiian2017unified,carlucci2018agnostic,akuzawa2019adversarial,ge2021supervised,nguyen2021domain,rahman2021discriminative,han2021learning}. This regularization ensures that the models extract features that are common to multiple domains and are therefore more likely to generalize to unseen domains. By mitigating the influence of domain-specific features, the idea is to improve the generalization capability of these models and ensure that they perform well on a variety of different domains.

In this paper, we build upon this line of research by investigating methods for learning domain-invariant features in machine learning models. Our proposed method is inspired by a simple intuition: while an image tends to convey rich but sometimes excessive details through its pixels, a corresponding text description can describe the crux of the image content in a highly concise and complementary manner; see Figure~\ref{fig:open}. Therefore, the most effective regularization might involve incorporating a regularization strategy in which the learned representations need to be close to the representations obtained from encoding the corresponding concise text descriptions of an image.

Building upon this argument, we propose a novel domain generalization approach that leverages recent advances in vision-language models, such as CLIP~\cite{radford2021learning}, to train our domain generalization models. We are particularly interested in the setting where our final models are relatively small, and thus, can benefit from a large pre-trained vision-language teacher model through distillation.  Our method, dubbed \name{} (Regularized Invariance with Semantic Embeddings), incorporates both the vision and language components of a pre-trained and frozen CLIP teacher, inspired by the importance of the representations encoded by the language component. Specifically, \name{} includes three loss functions: the empiricial risk minimization (ERM) loss that follows the standard pipeline of domain generalization, the model distillation loss that leverages the pretrained weights of the image component of CLIP, and the cross-domain (text to image) distance loss that uses the power of text through the language component of CLIP.

To fully harness the power of language, we introduce two different designs of the cross-domain distance loss function: the absolute distance design pushes the student's learned representation closer to the teacher's domain-invariant representation learned from language, while the relative distance design enforces that the relative domain distances in the teacher's encoded language space are transferred over to the learned representation in the student's encoded image space. 


\vspace{-10pt}
\paragraph{Contributions.} In summary, our main contributions are:
\begin{itemize}[noitemsep]
    \item To the best of our knowledge, we are the first to leverage knowledge distillation using a large vision-language model as a teacher for domain generalization. 
    \item We propose to regularize the representation learned by the student through images to be closer to the ones from the teacher's text representation, as text 
    can be more concise and capture the semantic essence. 
    \item We propose two loss functions, namely the absolute distance and the relative distance, which provide specific guidance on how the student model's training process should be regularized. 
    \item We conduct a rich set of experiments to validate the effectiveness of our model \name{} on domain generalization benchmarks and ablate the performance of each of its components. 
\end{itemize}

\section{Related Work}
\label{sec:related}
\subsection{Domain Generalization}

Domain Generalization \cite{muandet2013domain} has been widely studied in recent years.
It mainly studies the problem of training a model from the data collected from multiple 
source distributions and then testing the trained model 
on a target distribution that is different from the training ones. Because of this problem formulation, 
a natural assumption to guide the development of the methods 
is that
if the model can learn a representation that is invariant across the multiple 
training domains, 
it will generalize well to the unseen test domain. 
A large number of methods have been invented
following this natural assumption, 
aiming to force the invariance across samples of training distributions, 
either through explicit regularization based methods \cite{li2018domain,li2018deep,zhao2020domain,wang2016select,motiian2017unified,carlucci2018agnostic,akuzawa2019adversarial,ge2021supervised,nguyen2021domain,rahman2021discriminative,han2021learning,zhu2022localized,ding2022domain,wang2022out,wu2022siamdoge,meng2022attention,lee2022cross,zhang2022mvdg,yao2022pcl,huang2020improving,wang2022toward}
or (virtual) data augmentation methods \cite{shankar2018generalizing,yue2019domain,gong2019dlow,zhou2020deep,huang2021fsdr,wang2022domain}. 


In addition, 
the assumption above of ``invariance across multiple domains''
is being challenged in recent years 
with the argument that 
a more realistic scenario is when the training datasets are not necessarily 
partitioned into multiple distributions/domains with clear boundaries
during training. 
As a response to this argument, 
more powerful methods that do not rely on the domain partitions to force invariance
have been introduced
\cite{wang2020self,tian2022neuron,huang2022two}.
Our method in this paper in tested in the context of 
this challenging scenario. 

Also, in recent years, 
it seems the community is using the terminology 
\emph{out-of-distribution (OOD) generalization}
to largely refer to domain generalization. 
For more detailed discussions 
on the topics of domain generalization 
and out-of-distribution (OOD) generalization, 
we refer the reader to dedicated surveys
\cite{wang2022generalizing,shen2021towards}.

More closely related to our contribution in this paper, 
we notice a prior work that also leverages a pre-trained model to improve domain generalization performance.  Specifically, Domain Prompt Learning (DPL) \cite{zhang2021domain} utilizes a lightweight prompt adaptor to automatically generate a prompt that estimates domain-specific features given unlabeled examples from each distribution.
The following two works are not closely related to CLIP but leverage CLIP as their pre-trained model for domain generalization: \cite{li2022domain} dispatches proper pre-trained models (including CLIP) to each sample based on their generalization ability. \cite{cha2022domain} re-formulates the DG objective by mutual information with oracle models (including CLIP).

\emph{Key novelty:} Unlike prior work, we leverage CLIP as a teacher and regularize the student's learned representation through images to be closer to the corresponding text representation of the teacher. Our method includes two loss functions that directly leverage language for learning invariant image representations. 

\subsection{Knowledge Distillation}
Knowledge distillation is a technique for transferring knowledge from a teacher model to a student model, 
by optimizing the student model to match the outputs or intermediate features of the teacher model. 
This technique is used in numerous distinct contexts, such as semi-supervised learning \cite{orbes2019knowledge} or even self-supervised learning \cite{xu2020knowledge}.

Ever since the introduction of the term in 
\cite{DBLP:journals/corr/HintonVD15}, 
a plethora of techniques have been developed \cite{gou2021knowledge}
with improvement in various aspects, 
centering around the idea of 
how to align the output of the student model 
to the teacher model for every input, 
where the alignment and output are both subject to various concrete definitions. 
For example, 
one branch of works varies on how to enforce the alignment, 
with a particular focus on the loss function design 
over the outputs between the teacher and the student for every sample, 
with popular studies such as $\ell_1$ \cite{DBLP:conf/nips/KimPK18}, $\ell_2$ \cite{DBLP:journals/tnn/ChenWXXT21,DBLP:conf/aaai/PassbanWRL21,DBLP:conf/eccv/WangFLWLM20}, 
MMD \cite{DBLP:journals/corr/HuangW17a}, 
KL divergence \cite{DBLP:conf/aaai/ChenWZ18,DBLP:journals/tnn/PassalisTT21,DBLP:conf/cvpr/PassalisTT20}, 
and cross-entropy losses \cite{DBLP:conf/eccv/XuRLG20,DBLP:conf/cvpr/LiuWGTCOK19}.
Another branch studies how to define the output, which, 
at a high-level, 
has variants of directly using the embeddings from a certain (or final) layer \cite{DBLP:journals/corr/abs-2003-04289,DBLP:conf/eccv/GuanZWZYBT20,DBLP:conf/iccv/HeoKYPK019,DBLP:conf/aaai/ShenWSSS19}, 
or some more structured functions of the (pair-wise) embeddings of those layers \cite{DBLP:conf/fgr/LiPZ21,DBLP:conf/ijcai/ZhangP18a,DBLP:conf/cvpr/YimJBK17}.
There are also other branches 
such as the student-teacher architecture design 
or distillation algorithms that are not directly related to our study in this paper; 
we recommend the reader to refer to a survey for more details \cite{gou2021knowledge}. 

Among these works, the most relevant technical development to our method 
is to distill the preservation of the relationship between samples
from the teacher model to the student model. 
For example, 
\cite{DBLP:journals/tnn/ChenWXXT21} distills while the 
the student also learns the relationship between samples 
after the relationship is projected to a lower dimensional space, 
and other works more directly optimize
the similarity of 
the pair-wise distances between 
embeddings after each pair of samples 
is fed into the teacher and student models, respectively
\cite{DBLP:conf/cvpr/0004YLWCR19,DBLP:conf/iccv/PengJLZWLZ019,DBLP:conf/cvpr/LiuCLYHLD19,DBLP:conf/cvpr/ParkKLC19}. 

\emph{Key novelty:} 
The objective of prior work KDDG~\cite{wang2021embracing} is to distill the knowledge of a \emph{pure vision teacher} to a student model. In contrast, our approach focuses on distilling the knowledge of large-scale \emph{vision and language} models (CLIP) to the student model.



\subsection{Large Vision-Language Models}
Recent advances in vision-language models~\cite{radford2021learning,jia2021scaling,yang2022unicl,yao2021filip,yang2022vision,you2022learning} have shown promising results in learning generic visual representations and facilitating zero-shot transfer to diverse downstream classification tasks through the use of prompts. These models typically rely on a contrastive loss to align a visual encoder and a text encoder in a shared feature space. Trained on large-scale image-text pairs, these vision-language models demonstrate transferability across a wide range of applications, including object detection~\cite{gu2021open}, semantic segmentation~\cite{zhou2022extract}, and point cloud classification~\cite{zhang2022pointclip}. 

In particular, Contrastive Language Image Pre-training \ie, CLIP~\cite{radford2021learning} utilizes 400M pretraining image-text pairs to conduct image-caption contrastive pretraining.
Empirically, CLIP shows superior zero-shot image classification performance, achieving 76.2\% top-1  accuracy on the ImageNet validation set, which is on par with the performance of an ImageNet fully-supervised ResNet101 model. Furthermore, CLIP shows potential domain generalization capabilities. For example, it achieves 60.2\% accuracy on ImageNet Sketch Dataset while the ImageNet supervised training model~(ResNet101) can only achieve 25.2\% accuracy. This motivates us to answer the following question: \emph{What is the best way to distill CLIP's rich domain knowledge to a smaller student network for domain generalization tasks?}
\section{RISE: Regularized Invariance with Semantic Embeddings}
\label{sec:method}

\begin{figure*}[t]
\centering
\includegraphics[width=17cm,height=9cm,keepaspectratio]{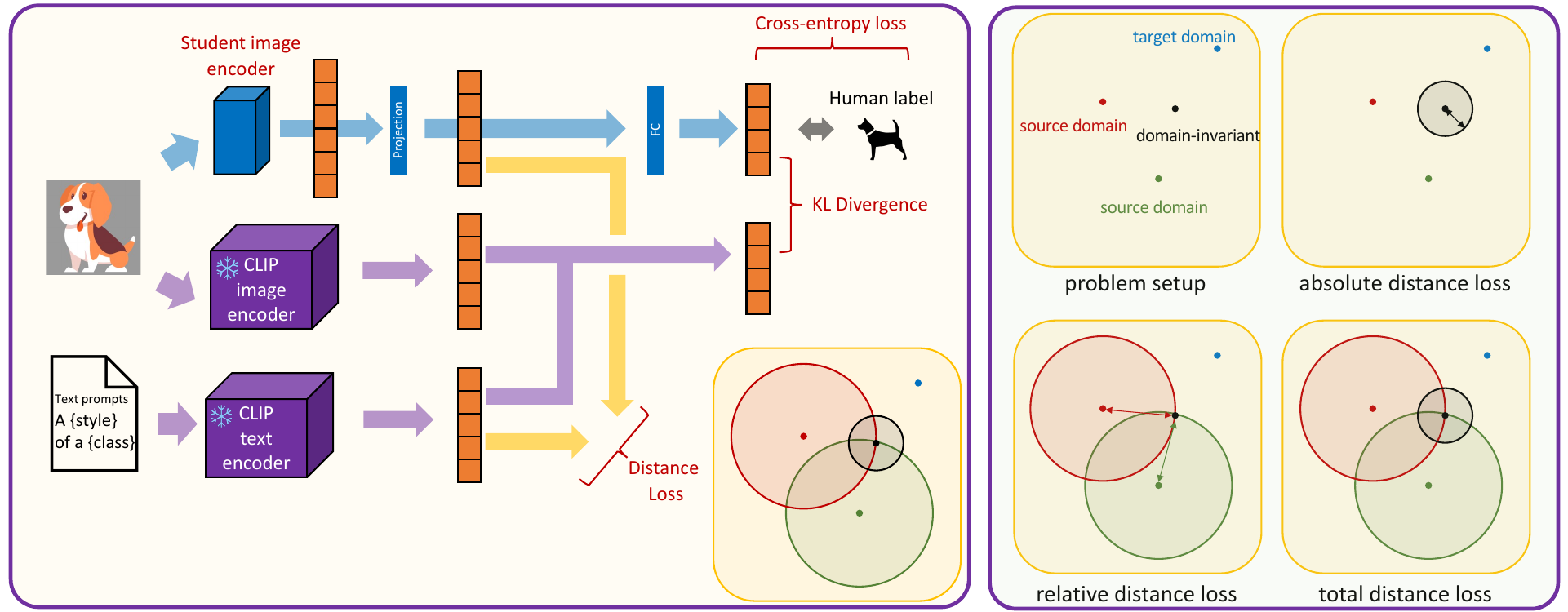}
\caption{\textbf{(Left)} Overview of the pipeline of our proposed method
as a combination of three losses: the cross-entropy loss as in standard supervised training, the KL divergence loss as in domain distillation, and our proposed cross-domain (text to image) distance loss. Here, a \emph{pre-trained and frozen} CLIP is the teacher model.  The teacher model is not trained.
\textbf{(Right)} The intuition of our two proposed losses and their combined effects. (i) In latent space, 
we aim to regularize the model to learn a representation that is close to the domain-invariant representation from the teacher's text space; (ii) the absolute distance loss can regularize the search to be within the shaded area; (iii) the relative distance loss can regularize the search to be within the overlap area; (iv) the combined loss can shrink the search space by overlapping these two. }
\label{fig:main}
\end{figure*}

In this section, we present the details of our approach for distilling a large vision-language model's learned semantic knowledge into a smaller student model for domain generalization.  Importantly, we use a \emph{pre-trained and frozen} CLIP~\cite{radford2021learning} as the teacher in this work.

\subsection{Notations, Baseline, and Distillation from Teacher's Image Component}

We first introduce our notations. 
We use $(\X, \Y)$ to denote the training dataset with $n$ (data,label) paired samples. 
these data samples can be from multiple domains or distributions, 
but 
since our model does not need the domain-ID information, 
we do not need a notation to specify which distribution the samples are from. 
Let $(\x,\y)$ denote one sample and $f(\cdot;\theta)$ denote the model we aim to train.  Thus, a vanilla paradigm of training a domain generalization 
model without domain IDs is as simple as the standard empirical risk minimization (ERM):
\begin{align}
    \sum_{(\x,\y)\in(\X,\Y)}l(f(\x;\theta), \y),
    \label{eq:erm}
\end{align}
where $l(\cdot,\cdot)$ denotes a generic loss function. 

We aim to incorporate rich prior knowledge from a CLIP pretrained image model teacher
through distillation. 
We use $h(\cdot;\phi)$ to denote this pretrained model, 
and the training process as:
\begin{align}
    \sum_{(\x,\y)\in(\X,\Y)} l(f_l(\x;\theta), h_l(\x;\phi)), 
    \label{eq:distill}
\end{align}
where $l(\cdot,\cdot)$ denotes a typical loss function that measures the distance of two vectors (e.g., KL divergence between the two predicted output distributions). 
$f_l(\cdot;\theta)$ and $h_l(\cdot;\phi)$ denote the output of logits instead of the final prediction. 


To use CLIP to produce output distributions in a classification setting, 
we feed in the text encoder of CLIP with queries 
generated with the label of the images (one query per label) 
such as 
``a photo of a dog'', ``a photo of an elephant'', etc. 
We use the image encoder's embedding to match each of the class query embeddings of the text encoder using cosine similiarity, and normalize the result to generate the output logits. 

\subsection{Regularization with Teacher's Language Component}

We use $g_l(\cdot;\delta)$ to denote the CLIP teacher's language component
that takes the input text phrase 
and generates an embedding. 
In general, if we have a generic description 
of the image, such as ``$\z=\textnormal{a photo of a dog}$'', 
we can directly feed this text phrase into the model to generate 
the corresponding
embedding, 
following $\mathbf{e_\z}(\textnormal{dog}) = g_l(\z;\delta)$. 

However, in practice, 
although ``a photo of a dog'' 
is recommended by CLIP as a standard text template, 
this text 
might not be generic enough as it still
indicates the pixel statistics of the image 
following the typical statistics of what a \emph{photo} has, which is 
potentially different from what a \emph{sketch} or what a \emph{painting} has. 

To overcome the potential over-reliance on the pixel statistics 
of \emph{photo}, we use the recommended list of eighty templates 
of text prompts by CLIP~\cite{radford2021learning}, including from ``a photo of my \{\}'', 
``an art of \{\}'', to even ``a tattoo of the \{\}''
and consider their averaged representation as 
the generic teacher's text representation 
of the depicted object. 

More concretely, we build the generic representation of class $i$ by 
\begin{align*}
    \mathbf{e_\z}(i) = \frac{1}{n}\sum_{\z \in \mathbf{Z}(i)}g_l(\z;\delta)
\end{align*}
where $\mathbf{Z}(i)$ denotes the set of recommended 
text templates when the class is filled in with the class
name corresponding to object $i$. 

With the teacher's generic text embedding $\mathbf{e_\z}(i)$, we aim to regularize the learning process of the student model to match its learned image representation to this generic representation, with two losses that function differently: absolute distance loss and relative distance loss. 

\subsubsection{Absolute Distance Loss}

The absolute distance loss is designed to directly push 
the student's learned image representation to be close to the teacher's generic text representation:
\begin{align}
    \sum_{(\x,\y)\in(\X,\Y)}\sum_{i\in\mathbf{I}} \mathbb{I}[c(\x)=i]k(f_l(\x;\theta), \mathbf{e_\z}(i)), 
    \label{eq:abs}
\end{align}
where $k(\cdot,\cdot)$ is a distance metric, $\mathbf{I}$ is the collection of all possible classes, and $\mathbb{I}[c(\x)=i]$ is simply an identity function that returns $1$ if the class of $\x$ is $i$ and $0$ otherwise. 

Ideally, if we can achieve the minimum value from \eqref{eq:abs}, we will train a student model that can learn generic 
visual representations that are likely to be invariant across the input 
domains and perform well on target domains. 

However, in practice, due to the difficulties of optimizing deep learning models on real-world data, the optimization cannot easily find such optimal solutions. Therefore, we need to introduce an 
additional regularization to help shrink the search space. 

\subsubsection{Relative Distance Loss}
We introduce a relative distance loss that can better describe where the target generic representation is. 

To do so, we need to first introduce several additional anchor points. For a domain generalization problem with possible training domain $d\in\mathbf{D}$, 
and for every class $i\in\mathbf{I}$, 
we generate $\mathbf{e}_\z(d, i)$ 
by feeding the text prompt 
``a \{d\} of $\{i\}$'' to the teacher's text encoder. 

Therefore, we have the relative distance loss as 
\begin{equation}
\begin{aligned}
    \sum_{(\x,\y)\in(\X,\Y)}&\sum_{i\in\mathbf{I}} \mathbb{I}[c(\x)=i]\sum_{d\in\mathbf{D}}\\
    &k_1\Big(k_2\big(f_l(\x;\theta), \mathbf{e}_\z(d, i)\big),
    k_2\big(\mathbf{e_\z}(i), \mathbf{e}_\z(d, i)\big)\Big), 
    \label{eq:rea}
\end{aligned}
\end{equation}
where $k_1$ and $k_2$ denote two distance metrics. 

Intuitively, the relative distance loss helps to pinpoint the location 
of the teacher's generic text representation by pushing the relative position 
of the student's learned representation from images with respect to those anchor points to be the same as the position of the generic representation with respect to the anchor points. 

The idea of the relative distance loss is to help the model to get to the generic embedding more directly. 
How it can help in searching for the generic representation is illustrated in the right-hand side of Figure~\ref{fig:main}.


\subsection{Full Method}

Connecting all the pieces above, our full method is to 
train the model with the loss functions from 
\eqref{eq:erm} to \eqref{eq:rea}, 
with hyperparameters to balance the contribution of each method; see the left-hand side of Figure~\ref{fig:main}.  

Formally, our final method is to train the model 
with 
\begin{align*}
    &\sum_{(\x,\y)\in(\X,\Y)} \lambda_1 l(f(\x;\theta), \y) 
    + \lambda_2 l'(f_l(\x;\theta), h_l(\x;\phi)) + \\
    & \sum_{i\in\mathbf{I}}\mathbb{I}[c(\x)=i]
    \lambda_3 \bigg(
     k(f_l(\x;\theta), \mathbf{e_\z}(i)) + \\
    & 
     \sum_{d\in\mathbf{D}}
    k_1\Big(k_2\big(f_l(\x;\theta), \mathbf{e}_\z(d, i)\big),
    k_2\big(\mathbf{e_\z}(i), \mathbf{e}_\z(d, i)\big)\Big)
    \bigg)
\end{align*}
where $\lambda_1$, $\lambda_2$, and $\lambda_3$
are three hyperparameters that balance each loss term. 

\subsection{Implementation Details}

In practice, we implement $l$ as cross-entropy loss, 
$l'$ as KL divergence, $k$ as CosineSimilarity, $k_1$ as CosineSimilarity, and $k_2$ as L2 loss. The KL divergence $l'$ introduces one more hyperparameter temperature $t$ to control the smoothness of CLIP's predictions. The detail of distance metrics selection is analyzed in the ablation study.
We use one linear layer to project the image embedding of the student to the text embedding of the CLIP teacher. During inference, images are passed through the student image encoder and FC layer (or CLIP's text embedding) to make final predictions.

\section{Experiments}
\label{sec:exp}
We evaluate our approach in leveraging language as a regularization strategy for training a student image model for domain generalization.  We compare state-of-the-art domain generalization methods and perform ablation studies to analyze the various components of our model.  

\subsection{Setup}
We follow the setting in \cite{gulrajani2020search, ye2021ood} and evaluate our domain generalization approach. Specifically, we use the same strategy for model selection, dataset splitting, and network backbone. 




\begin{table*}[h]
\caption{Results of domain generalization methods with ResNet backbone. Ens/MA stands for Ensemble/ Moving Average. $\star$ denotes fine-tuning on target datasets. Hint~\cite{hinton2015distilling} stands for the distillation loss. AD stands for absolute distance loss. RD stands for relative distance loss. MT stands for Mix Teacher engineering technique. We report averaged accuracy across three runs.} 
\vspace{5pt}
\label{table:DGmethods} 
\footnotesize
\centering 
\begin{tabular}{l|c | c | c c c c | c} 
\hline 
Method& Backbone & Ens/MA & PACS & VLCS & OfficeHome & Terra  & Ave\\ [0.5ex] 
\hline \hline
ERM  \cite{ye2022ood}& ResNet18 & No & 81.5 & 73.2 & 63.3 & 43.6  & 65.4\\
Best SoTA competitor& ResNet18 &No & 83.4~\cite{huang2022two} & 74.1~\cite{huang2022two} & 63.8~\cite{li2018domain} & 44.5~\cite{huang2022two} & 66.5  \\
\hline
ERM ~\cite{gulrajani2020search}& ResNet50 & No & 85.7 & 77.4 & 67.5 & 47.2  &69.5 \\ 
Best SOTA competitor & ResNet50 & No & 86.6 \cite{seo2020learning}& 78.8~\cite{sun2016deep} & 68.7~\cite{sun2016deep} & 48.6 \cite{nam2021reducing} & 70.7\\
\hline
Ensemble \cite{arpit2021ensemble}& ResNet50 & Yes & 87.6 & 78.5 & 70.8 & 49.2 & 71.5 \\
SWAD \cite{cha2021swad} & ResNet50 & Yes & 88.1 & 79.1 & 70.6 & 50.0 & 71.9 \\
EoA \cite{arpit2021ensemble} & ResNet50  & Yes & 88.6  & 79.1 & 72.5 & 52.3 & 73.1 \\
\hline \hline
CLIP \cite{zhang2021domain} (Teacher) & ViT B/16 & No &  96.1 & 82.3 & 82.3 & 50.2$\star$ & 77.7  \\
\hline
ERM + Hint & ResNet18 & No & 84.6 & 78.0 & 64.6 & 47.0 & 68.6 \\
ERM + Hint + AD       & ResNet18 & No & 85.1 & 78.5 & 65.6 & 48.2 &  69.4 \\
ERM + Hint + RD       & ResNet18 & No & 84.9 & 78.2 & 65.2 & 47.9 &  69.0\\
\rowcolor{lightgray!30}ERM + Hint + AD + RD (Our full method) & ResNet18 & No & 85.3 & 78.6  & 65.9 & 48.4 &  69.6\\
\hline
ERM + Hint & ResNet50& No & 88.4 & 80.7 & 70.2 & 50.5 &  72.5\\
ERM + Hint + AD      & ResNet50& No & 89.0 & 81.5 & 71.3 & 52.2 &  73.5\\
ERM + Hint + RD      & ResNet50& No & 88.8 & 81.2 & 71.1 & 51.7 &  73.2\\
\rowcolor{lightgray!30}ERM + Hint + AD + RD (Our full method) & ResNet50& No & 89.4 & 81.7 & 71.6 & 52.3 &   73.8\\
\rowcolor{lightgray!30}ERM + Hint + AD + RD + MT (Our full method) & ResNet50  & Yes & 90.2 & 82.4 & 72.6 & 54.0 &  74.8 \\
\hline
\end{tabular}
\end{table*}

\subsection{Datasets, Hyperparameter Search, and Model
Selection}

We follow DomainBed~\cite{gulrajani2020search} and Ood-bench~\cite{ye2021ood} to choose datasets that cover as much variety as possible from the various OOD research areas for our experiments. We conduct experiments on four OOD datasets:  Terra Incognita~\cite{beery2018terra}, OfficeHome~\cite{venkateswara2017officehome}, VLCS~\cite{torralba2011vlcs}, and PACS~\cite{li2017pacs}.

To be consistent with the existing line of work, we use the training-validation protocol for model selection: given $N$ domains, it uses 90\% the amount of data in $N-1$ domains for training, the other 10\% for validation, selects the best model based on the validation result, tests the model on the held-out domain and reports this result. 

There are altogether four hyperparameters for our method -- the weights of supervised loss $\lambda_1$, distillation loss $\lambda_2$, distance losses $\lambda_3$, and temperature $t$. Overall, we set the hyperparameter search space of our method as $\lambda_1 \in [0.1, 1.0]$, $ \lambda_2 \in [0.1, 1.0]$, $\lambda_3 \in [0.1, 1.0]$, $t \in [1.0, 3.0]$.
We adopted the same hyperparameter search protocol used in~\cite{gulrajani2020search,ye2022ood}.

\subsection{Empirical Results}

\paragraph{Zero-shot performance of CLIP teacher.}
We select CLIP ViT-B/16 as the teacher for the following experiments (due to limited computational resources we could not try larger models). In Table~\ref{table:DGmethods}, CLIP ViT-B/16 achieves 96.1\%, 82.3\%, 82.3\%, and 34.1\% on PACS, VLCS, Office-Home, and Terra Incognita respectively when performing zero-shot inference. Except for Terra Incognita, both CLIP models outperform the best individual state-of-the-art results by up to 7\%. 
Because of the extremely low zero-shot accuracy on Terra, we finetune CLIP on Terra to obtain a better CLIP teacher, which achieves 50.2\% zero-shot accuracy.
Overall, we use finetuned CLIP ViT B/16 for Terra Incognita and zero-shot CLIP ViT B/16 for the remaining three datasets.


\vspace{-10pt}
\paragraph{Comparison with existing DG methods} 
We compare to the recent top DG algorithms, using both ResNet18 and ResNet50 pre-trained on ImageNet-1k as the backbone. Our results are presented in Table~\ref{table:DGmethods}. The ``Best SoTA competitor" refers to the highest performance in the literature within the standard DG experimental protocol, and the numbers listed under this category may be from different methods. In addition, we also include ensemble and weight-averaging techniques in the third-row panel.

We first study the effect of the standard distillation loss~\cite{hinton2015distilling}. We use the soft labels produced by the CLIP teacher as an additional target for the student to match, in addition to the (one-hot) ground-truth labels. This is done by minimizing the KL-divergence between the predicted distributions of the student and the teacher. Training a student with human labels and distillation loss (ERM + Hint in the last two panels), already outperforms most of the state-of-the-art methods on the benchmarks. EoA~\cite{arpit2021ensemble}, a moving average variant method, is the only method that outperforms ERM + Hint with the ResNet50 backbone.  
Next, we study the effect of our proposed method. We observe that adding absolute distance (AD) and relative distance (RD) to ERM + Hint both result in clear performance gains, and together produce the best results which indicate their complementarity. For ResNet 18, AD, RD, and AD + RD provide 0.8\%, 0.4\%, and 1.0\% improvement over ERM + Hint respectively. For ResNet 50, AD, RD, and AD + RD provide 1.0\%, 0.7\%, and 1.3\% improvement over ERM + Hint respectively.

 
\subsection{Ablation Studies}

In this section, we study the impact of each component in our method. We evaluate our method with a ResNet50 backbone on the most popular DG benchmark PACS to conduct the following experimental analyses.

\subsubsection{Impact of using text embedding as supervision}

\begin{table}[htbp]
\begin{center}
\footnotesize
\begin{tabular}{l | l |  c }
\hline
Method & Embedding  & Acc \\
\hline \hline
ERM + Hint + AD & Image  & 88.4\\
\rowcolor{lightgray!30}ERM + Hint + AD & Text  & 89.0\\
\hline
ERM + Hint + RD & Image  & 88.1\\
\rowcolor{lightgray!30}ERM + Hint + RD & Text  & 88.8\\
\hline
\end{tabular}
\vspace{5pt}
\caption{Analysis of using CLIP's image embedding and text embedding as supervision. Hint~\cite{hinton2015distilling} stands for the distillation loss. AD stands for absolute distance loss. RD stands for relative distance loss.}
\label{table:textregu}
\end{center}
\end{table}

We study the impact of using CLIP's text embedding and image embedding as supervision for our absolute distance loss and relative distance loss.
The results displayed in Table \ref{table:textregu} indicate that for both our absolute distance loss and relative distance loss, utilizing text embedding of CLIP as supervision yields better results compared to using the image embedding counterpart for regulating the learning process of our model, despite having the same loss function setting. Specifically, ERM + Hint + AD and ERM + Hint + RD with text embedding supervision outperform their image embedding counterparts with 0.6\% and 0.7\% improvement respectively. This analysis helps validate our assumption that CLIP's text embedding contains rich semantic information, and it can be treated as a domain-invariant representation since it is independent of images.
In addition, Table \ref{table:textregu} demonstrates that both our absolute distance loss and relative distance loss exhibit comparable performance under the ERM + Hint setting. Specifically, ERM + Hint achieved 88.4\% (+AD) and 88.1\% (+RD) using image embedding. Under ERM + Hint setting with text embedding, absolute distance loss performs slightly better than relative distance loss where ERM + Hint attains 89.0\% (+AD) and 88.8\% (+RD).

\subsubsection{Impact of each loss component}

\begin{table}[htbp]
\begin{center}
\footnotesize
\begin{tabular}{l | c }
\hline
Method  & Acc \\
\hline \hline
ERM & 85.7 \\
ERM + Hint &  88.4\\
ERM + AD & 87.8 \\
ERM + RD &  87.2\\
ERM + Hint + AD &  89.0\\
ERM + Hint + RD &  88.8\\
\rowcolor{lightgray!30}ERM + Hint + AD + RD & 89.4\\
\hline
\end{tabular}
\vspace{5pt}
\caption{Analysis of the effectiveness of each loss function in our method using ResNet50 backbone on PACS. Hint~\cite{hinton2015distilling} stands for the distillation loss. AD stands for absolute distance loss. RD stands for relative distance loss.}
\label{table:eachloss}
\end{center}
\end{table}

Table \ref{table:eachloss} demonstrates that each component of our loss function contributes to the final performance. By adding one additional loss component to ERM (85.7\%), ERM + Hint (88.4\%), ERM + absolute (87.8\%), and ERM + relative (87.2\%), all get substantial improvements: +2.7\%, 2.1\%, 1.5\%, respectively. Interestingly, ERM + AD achieves comparable performance with ERM + Hint which suggests that using CLIP's text embedding as supervision has the potential to match the performance of using the entire CLIP model. That is, a CLIP teacher can be used to generate supervisory signals for distillation without having access to any images. 
Moreover, by adding absolute distance loss and relative distance loss to ERM + Hint, there are further improvements of 0.6\% and 0.4\%, respectively, for ERM + Hint + AD and ERM + Hint + RD. Finally, by combining all components and using ERM + Hint + AD + RD (89.4\%), we observe a significant improvement of 3.7\% compared to using ERM only (85.7\%).

\subsubsection{Impact of prompt engineering and ensemble}

\begin{table}[htbp]
\begin{center}
\footnotesize
\begin{tabular}{l |l |c }
\hline
Method & Template  & Acc \\
\hline \hline
ERM + Hint + AD & a photo of a $\{class\}$ & 88.5 \\
\rowcolor{lightgray!30}ERM + Hint + AD & Ensemble template & 89.0 \\
\hline
ERM + Hint + RD & a photo of a $\{class\}$ & 88.3 \\
\rowcolor{lightgray!30}ERM + Hint + RD & Ensemble template &  88.8\\
\hline
\end{tabular}
\vspace{5pt}
\caption{Analysis of CLIP's prompt engineering end ensemble. Hint~\cite{hinton2015distilling} stands for the distillation loss. AD stands for absolute distance loss. RD stands for relative distance loss.}
\label{table:prompt}
\end{center}
\end{table}
Table \ref{table:prompt} demonstrates the effectiveness of having a prompt ensemble template, which enhances the accuracy compared to a single prompt template. Both ERM + Hint + AD and ERM + Hint + RD settings display an accuracy improvement of 0.5\%. The ensemble template utilizes 80 representative templates of text prompts by CLIP~\cite{radford2021learning}. The improvement in accuracy suggests that the text embedding generated by the ensemble template is more robust than the single template counterpart (i.e., ``a photo of a \{\}'') when facing distribution shift tasks. 
For those interested in exploring the details of the eighty prompt ensemble templates, they can be found \href{https://github.com/openai/CLIP/blob/main/notebooks/Prompt_Engineering_for_ImageNet.ipynb}{here}.

\subsubsection{Impact of different distance metrics}

\begin{table}[htbp]
\begin{center}
\footnotesize
\begin{tabular}{c |  c |c }
\hline
Method & Loss & Acc \\
\hline \hline
\rowcolor{lightgray!30}ERM + Hint + AD & CosineSimilarity & 89.0\\
ERM + Hint + AD & Supervised Contrastive & 88.6\\
ERM + Hint + AD & L1 & 88.0\\
ERM + Hint + AD & L2 & 88.1\\
\hline
ERM + Hint + RD & KL & 88.7\\
ERM + Hint + RD & L1 & 88.3\\
\rowcolor{lightgray!30}ERM + Hint + RD & L2 & 88.8\\
\hline 
\end{tabular}
\vspace{5pt}
\caption{Effect of different regularization and distance metrics. For ERM + Hint + RD, we fix $k_2$ to be consine similarity, and only explore which kind of distance metric $k_1$ works the best with $k_2$. Hint~\cite{hinton2015distilling} stands for the distillation loss. AD stands for absolute distance loss. RD stands for relative distance loss.}
\label{table:distance}
\end{center}
\end{table}

Table \ref{table:distance} shows the variation in performance due to different regularization and distance metrics. When considering the ERM + Hint + absolute distance loss setting (i.e., choosing the distance metric $k$ in Eqn.~\ref{eq:abs}), the Cosine Similarity loss (89.0\%) outperforms the Supervised Contrastive (88.6\%), L1 (88.0\%), and L2 (88.1\%) approaches. (For Supervised Contrastive, the positive/negative pairs are the student's image feature and teacher's text feature for the ground-truth/non-ground-truth class.)  On the other hand, for the ERM + Hint + relative distance loss setting (i.e., choosing the distance metric $k_1$ in Eqn.~\ref{eq:rea}; to be consistent with the final absolute distance metric, we set $k_2$ to be cosine similarity), KL (88.7\%) and L2 (88.8\%) exhibit similar performance and outperform L1 (88.3\%). Overall, we implement both $k$ and $k_2$ as ConsineSimilarity and $k_1$ L2 distance metrics. 

\subsubsection{Impact of Mix Teacher}

\begin{table}[htbp]
\begin{center}
\footnotesize
\begin{tabular}{c |  c | c|c|c|c|c }
\hline
 Teacher & Ensemble & A & C & P & S & Avg \\
\hline \hline
CLIP ViT B/16 & No & 88.0 & 85.2 & 97.8 & 86.4 & 89.4\\
CLIP RN101 & No & 87.6 & 86.1 & 97.6 & 85.1 & 89.1\\
\rowcolor{lightgray!30}Mix Teacher & Yes & 88.7 & 86.7 & 98.3 & 86.9 & 90.2\\
CLIP ViT B/16 & Yes & 88.3 & 86.0 & 98.1 & 86.7 & 89.8\\

\hline 
\end{tabular}
\vspace{5pt}
\caption{Results of ERM + Hint + AD + RD with different CLIP teachers on PACS. Hint~\cite{hinton2015distilling} stands for the distillation loss. AD stands for absolute distance loss. RD stands for relative distance loss. MT stands for Mix Teacher engineering technique. A, C, P, and S: art-painting, cartoon, photo, and sketch.}
\label{table:mixteacher}
\end{center}
\end{table}

Finally, we explore the impact of having multiple CLIP teachers, which we call ``Mix Teacher'' (MT).  Specifically, we use another CLIP ResNet 101 as a teacher model, which achieves 94.9\%, 80.0\%, and 76.0\% zero-shot inference on PACS, VLCS, and Office-Home, respectively, and 50.5\% finetune inference on TerraIncognita.  Our ERM + Hint + AD + RD method with this CLIP RN101 teacher achieves 89.1\%, 81.6\%, 70.9\% and 52.3\% on PACS, VLCS, OfficeHome and Terra respectively. 

Table \ref{table:mixteacher} shows the ensembling results.  Overall, an ensemble of teachers achieves higher accuracy compared to non-ensemble teachers; from Table \ref{table:mixteacher}, we see that Mix Teacher of CLIP ViT B/16 + CLIP RN101 (90.2\%) exhibits better performance than CLIP ViT B/16 non-ensemble (89.4\%) and CLIP RN101 (89.1\%). We also investigate ensembling the outputs of two separate students trained with the same CLIP ViT B/16 teacher (Table \ref{table:mixteacher} last row). This ensemble model also does better (+0.4\%) than a single student (1st row), but not as well as ensembling multiple teachers.  

Although the overall performance is close between students distilled by different CLIP teachers (row 1 vs row 2 in Table \ref{table:mixteacher}), upon closer inspection, we find that the student distilled with CLIP VIT outperforms the CLIP RN counterpart on sketch domains and worse on cartoon domains. We suspect that the teacher CLIP with different model architectures have different domain biases and perform well on different domains. Thus, by ensembling student models distilled with different CLIP teachers that have different network architectures, we can further improve the generalization capability of our student method. We report the ensemble performance for two different ResNet50 pretrained student models with mixed teachers in the last row in Table~\ref{table:DGmethods}.  It provides a +1.0\% boost (74.8\% average accuracy) over our single non-ensemble model.





\section{Conclusion}
\label{sec:con}
One of the challenges in domain generalization is that machine learning models tend to learn domain-specific features, which can make them less effective at generalizing to new domains. This is because domain-specific features may be highly relevant to the training data but may not be useful for prediction on new domains.
To address this challenge, the community has focused on developing methods to regularize the learned representations to become domain-invariant features. 

In this paper, we build upon this direction by investigating methods for learning domain-invariant features in machine learning models. 
Our proposed method is inspired by the intuition that while an image tends to convey rich but sometimes excessive details through its pixels, a corresponding text description can describe the crux of the image content in a highly concise and complementary manner.

Following this intuition, we proposed \name{}, with main loss functions (the absolute distance and relative distance) to offer specific guidance on how the student model's training process should be regularized, that provides a powerful new direction for domain generalization research by incorporating the power of language to regularize image representations.

Our results suggest that leveraging language as a regularization strategy can achieve state-of-the-art performance on popular domain generalization benchmarks. We have demonstrated the effectiveness of our approach through a comprehensive set of experiments. 

In conclusion, \name{} provides a new direction for domain generalization research by incorporating the power of language to regularize image representations. Our results suggest that leveraging language as a regularization strategy can significantly improve the generalization capability of machine learning models, and we believe that our work can motivate further research in this direction.

\vspace{-10pt}
\paragraph{Limitations.} When facing the downstream task, such as Terra where CLIP shows poor performance during zero-shot inference, finetuning CLIP on the downstream task is recommended before distilling knowledge to students.

In addition, the quality and relevance of text descriptions used to regularize image representations may impact the effectiveness of our approach. In our experiments, we addressed this limitation by using an average of 80 different text descriptions for each image.  However, obtaining a more direct and generic text description might help improve the efficiency of the method.  We leave this to future work.

\vspace{-10pt}
\paragraph{Acknowledgements.}
This work was supported in part by NSF CAREER IIS2150012, NASA 80NSSC21K0295, and Institute of Information \& Communications Technology Planning \& Evaluation(IITP) grants funded by the Korean government (MSIT) (No. 2022-0-00871, Development of AI Autonomy and Knowledge Enhancement for AI Agent Collaboration) and (No. RS-2022-00187238, Development of Large Korean Language Model Technology for Efficient Pre-training.

{\small
\bibliographystyle{ieee_fullname}
\bibliography{egbib}
}

\end{document}